# Airlift Challenge: A Competition for Optimizing Cargo Delivery

**Adis Delanovic[1*], Carmen Chiu[1*], John F. Kolen[2], Marvin Gülhan[3], Jonathan Cawalla[3], Andre Beckus[1]**
[1] Air Force Research Lab, Information Directorate
[2] New Mexico Institute of Mining and Technology
[3] Fraunhofer FKIE
* Authors contributed equally to this work

**Abstract**

Airlift operations require the timely distribution of various cargo, much of which is time sensitive and valuable. These operations, however, have to contend with sudden disruptions from weather and malfunctions, requiring immediate rescheduling. The Airlift Challenge competition seeks possible solutions via a simulator that provides a simplified abstraction of the airlift problem. The simulator uses an OpenAI gym interface that allows participants to create an algorithm for planning agent actions. The algorithm is scored using a remote evaluator against scenarios of ever-increasing difficulty. The second iteration of the competition was underway from November 2023 to April 2024. This paper describes the competition, simulation environment, and results. As a step towards applying generalized planning techniques to the problem, a temporal PDDL domain is presented for the Pickup and Delivery Problem, a model which lies at the core of the Airlift Challenge.

## Introduction

An airlift operation comprises movement of cargo, supplies and/or personnel often within strict time limits. This is hampered by various factors, ranging from excessive amounts of cargo/personnel to conditions at the airports themselves, e.g., very limited space for taxiing or parking (Owen, 1997). Usually, the air cargo consists of time sensitive and valuable supplies. Even just a small change can result in a large amount of disruption, requiring extensive replanning which takes time and resources. This drives an increased need for automation within the planning process. The Airlift Challenge competition is intended to answer this need[1].

The airlift planning problem is difficult because of the number of constraints and factors. Often, Mixed Integer Linear Programming (MILP) solutions are utilized, but these require recalculation of the entire solution when a disruption occurs (Bertsimas, Chang, Mišić, & Mundru, 2019). However, reinforcement learning has also shown promise in quick replanning. Even if a solver can find an optimal solution to a problem, this solution must be created quickly in order to deal with dynamic elements.

The first iteration of the Airlift Challenge was held with the SPIE conference in 2023 and attracted 20 registered users and two high quality submissions that received 1st and 2nd place respectively (Delanovic, et al., 2023). The remainder of this paper describes the second iteration held in association with ICAPS 2024 and concluded in April 2024. It then presents a Pickup and Delivery PDDL (Planning Domain Definition Language) domain as a starting point for a potential submission to future IPCs (International Planning Competitions)[2].

## Simulation Environment Details

The goal of the competition's simulation environment is to provide a simplified model that captures the major factors influencing an airlift operation. This approach is based upon the Flatlands competition which was focused on train vehicle routing (Mohanty, et al., 2020). The Flatlands competition focused on "flattening", or simplifying, a complex environment into a set of the most important parameters. The advantage of a simplified environment is it abstracts away details in the environment, while the design of the starter kit allows many types of solutions to be used with it.

The model consists of an air network graph, where each node corresponds to an airport and each edge denotes the route that agents can traverse. The agents are airplanes that travel preset routes between airports. The agents move the same predetermined distance at every time step based on the speed of their airplane type. Airplanes are limited by fuel, routes available, maximum loading weight and speed. Airports themselves can only process a certain number of airplanes at a time, creating possible bottlenecks. A summary of the simulated objects and associated parameters is shown in Table 1Table *1*.

---

[1] See main website https://airliftchallenge.com/

[2] This paper is an extended version of one presented at the Workshop on the International Planning Competition at ICAPS 2024 (Delanovic, Chiu, & Beckus, Airlift Challenge: A Competition for Optimizing Cargo Delivery, 2024)

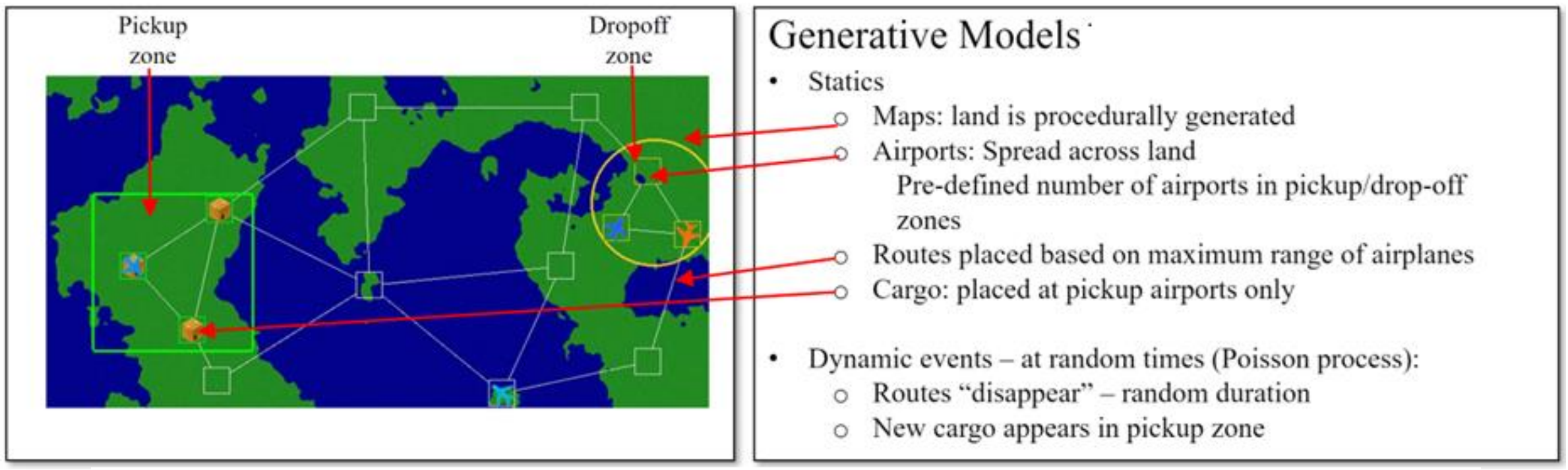

Figure 1: Simulator visualization highlighting aspects of generative models

| Object | Parameters |
|---|---|
| Airport | • Location<br>• Working capacity (# of airplanes that can process/load/unload at same time) |
| Route | • Flight Cost (determined based on distance)<br>• Flight Time (determined based on distance and flight speed)<br>• Start/end airports<br>• Availability status (routes may become unavailable at random intervals) |
| Airplane | • Cargo weight capacity<br>• Time to process<br>• Route Map: Contains all routes available to travel<br>• Processing priority (determines order in which planes process/load/unload) |
| Cargo | • Weight<br>• Source/destination airports<br>• Soft delivery deadline<br>• Hard delivery deadline |

Table 1: Model Parameters

At each time-step all agents receive an observation and corresponding reward. The agents then decide what action to take next using the policy provided by the competition participant, and pass it to the environment. This continues until either 1) all the cargo is delivered, or 2) a maximum number of time steps has elapsed. For the competition, each scenario is limited to 5,000 time steps.

An agent can be in one of four states at any given time: waiting, processing, ready-for-takeoff, or moving. An agent is put into a waiting state after landing at an airport. While in the waiting state it is held in a priority queue. Each airport has a finite processing capacity. When an airport has capacity and the airplane is next in the priority queue, it will transition into the processing state. While in this state, cargo loading or unloading actions will take place. After processing cargo, the airplane will then transition to the ready-for-takeoff state. An airplane will remain in the ready-for-takeoff state until it receives actions specifying its next destination. Then, the airplane will transition to the moving state, restarting the entire process over again.

Plans are implemented by issuing actions to change the state of an airplane. The actions and their associated preconditions and effects are briefly summarized in Table *2*. The *Process* action transitions the airplane from the waiting state into the processing state. If processing capacity is reached, the airplane waits in a priority queue based on a priority given by the agent policy. Processing must be performed at least once after landing, to represent re-fueling and other maintenance, even if there is no cargo to load/unload. Concurrent with processing, the *Load* and *Unload* actions may be used to transfer cargo. The *Takeoff* action receives a destination and transitions from ready-for-takeoff into a moving state. A valid destination is defined as any neighboring node of the current airport with respect to the route graph. Note that the simulator offers a consolidated action space which allows actions to be issued while in flight to the next airport, with sequencing at the airport handled automatically. More details are provided in the competition documentation[3].

| Action | Preconditions | Effects |
|---|---|---|
| *Process* | There is available capacity to process. | Sends airplane into processing state. |
| *Load* | New cargo will not exceed airplane capacity. Airplane is in processing state. | Loads a piece of cargo. |
| *Unload* | Airplane has cargo onboard and is in processing state. | Unloads a piece of cargo. |
| *Takeoff* | Airplane has finished processing. | Sends airplane into moving state. |

Table 2: Key components of action. All actions are undertaken while an airplane is landed at an airport.

[3] See https://airliftchallenge.com/chapters/ch2_model/main.html and https://airliftchallenge.com/chapters/ch3_interface/main.html for more details.

A malfunction generator will randomly cause disruptions to airplane routes, forcing the planner to reschedule. Malfunction are generated according to a Poisson distribution with the mean number of malfunctions per time step $\lambda$ predetermined before generating the scenario. Throughout each scenario, cargo will be dynamically generated at random times. In addition, the simulator has the option to place cargo and airplanes in random initial places.

Cargo is placed within a designated area called the "pick up zone". The pick-up zone contains a subset of airports. The cargo must be picked up and delivered to the "drop off zone" that is generally on the other side of the map. An illustration of these zones is shown in Figure 1. The drop-off and pick-up zones contain a subset of airports within a specific area.

Cargo is assigned a weight, a soft deadline and a hard deadline. The weight of the cargo varies. As shown in Figure *2*, cargo is considered to be on-time if it is delivered within the soft deadline. If the cargo is delivered past the soft deadline, it is considered late. If the agents are unable to deliver the cargo within the hard deadline, the cargo is considered missed, and the agent is penalized heavily for it.

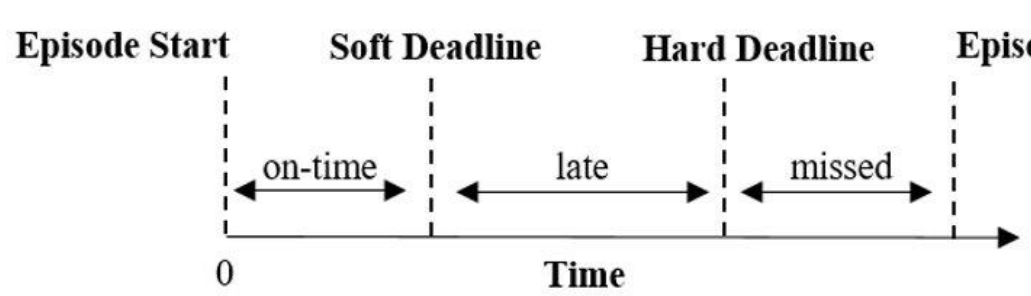

Figure 2: Deadlines for delivery.

Maps are generated using the Perlin noise algorithm, whereby a certain threshold value decides if an area is considered land or water, and airports are placed uniformly at random on the landmasses (Perlin, 1985). The generator also ensures that airports are not placed too close to each other. An example of a scenario can be seen in Figure 1.

The environment supports multiple airplane types, since this is required for successful airlift operations given the variety of airplanes and potential landing areas. Airplane types have speed, maximum travel range and maximum carrying capacity. The edges on the graph network correspond to the maximum range of the airplane type. The disparate graph networks for each airplane type are combined to from a single muti-graph where the edges contain data on which airplane type can traverse it. Certain airplane types may not be able to reach all airports and would have to work together with the others to complete cargo delivery.

For the ICAPS Airlift Challenge, agent priority was introduced. Given the finite processing capacity of airports, this addition enables solutions to prioritize agents for processing, regardless of their arrival time. Proper prioritization of agents provides the ability to improve cargo delivery outcomes from missing the hard deadline to late or even on-time.

[4] See https://codalab.lisn.upsaclay.fr/competitions/16103

## Baselines

There are three baselines provided. These are the random agent, shortest paths as well as a MILP. The random agent samples the observation and generates a random valid action. The shortest paths agent only traverses airports using the shortest path available. If a route along a path becomes unavailable, the agent will re-route to the next shortest path.

The MILP solution uses constraints as formulated in (Bertsimas, Chang, Mišić, & Mundru, 2019). It utilizes the GNU Linear Programming Kit (GLPK), an open-source solver that can be used for linear programming or MILPs (Makhorin, 2012). This solution does not scale well when additional agents are added to the scenarios and currently only runs with one agent and simple scenarios.

## Scoring

The competition evaluator is hosted on Amazon Web Services (AWS). The services do not allow communication with any outside resources. This ensures that all participants receive an even playing field and that there is no bias in scoring the submission. A participant will use the CodaLab platform to make a submission (Pavao, et al., 2022)[4]. The submission is immediately sent to AWS, where it will start the evaluation process. Results are shown on a leaderboard.

For each phase of the competition, over four hundred scenarios are generated to score the submitted algorithms. Scenarios increase in difficulty. This is accomplished by changing "static" as well as "dynamic" parameters. The static parameters include the quantity of agents, airports and cargo. The dynamic parameters include the rate at which new cargo is generated, as well as how often routes become unavailable. The evaluation scenarios are not released to the public, although a test set is provided.

A submission will run until it either reaches the allotted four-hour time limit, exceeds an overall 30% missed deliveries, or completes all scenarios. Upon either of these occurring, a score is generated and placed on the leaderboard. Figure 3 shows visualizations of the environment for four tests, ranging from the simplest to most difficult scenarios. The blue and white colored edges represent different airplane types.

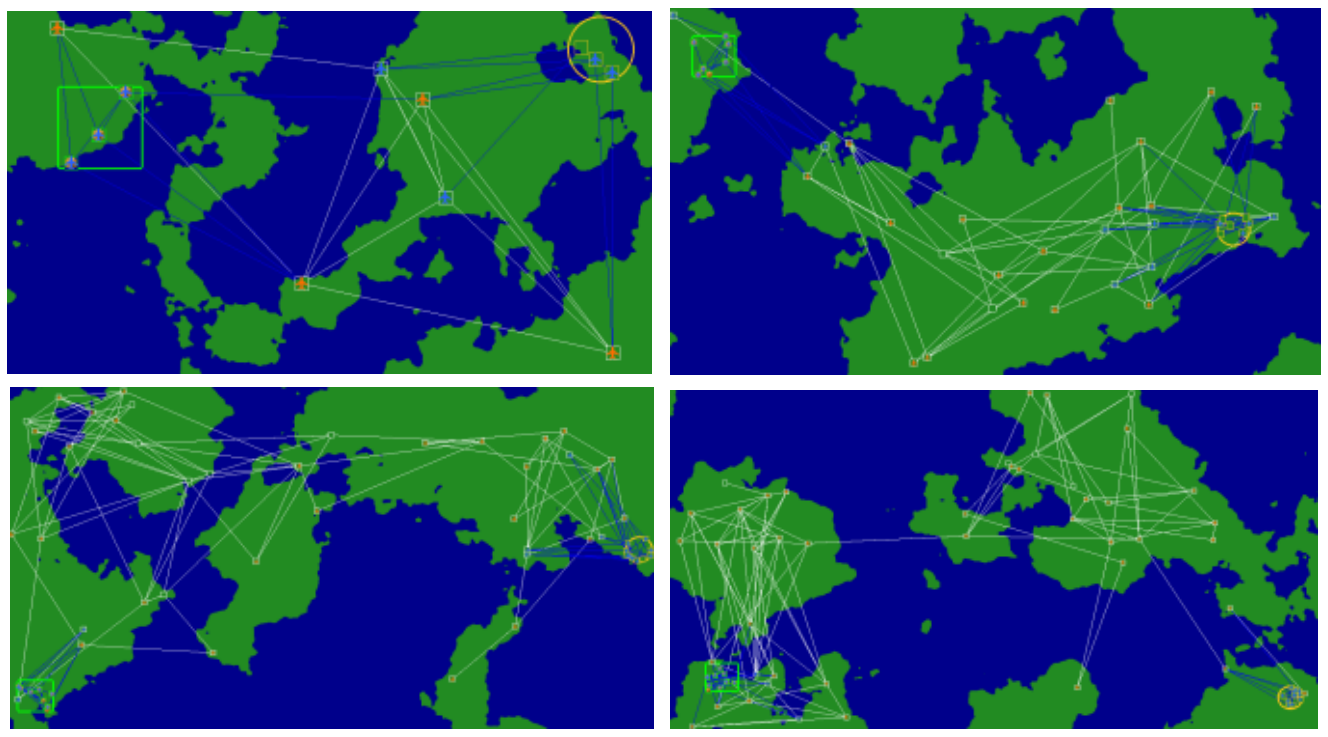

*Figure 3 - Example scenarios exhibiting difficulty increase.*

To generate an overall score for any submission, an episode score is assigned. For each level that the submission completes, a score is assigned based on missed cargo, late cargo and total flight cost. An overall score is calculated by summing the normalized score over all tests and levels.

A starter-kit is provided on the competition GitHub page with detailed description on what a submission should include[5]. Additionally, instructions are included on generating custom-made scenarios, and the competition organizers provide a large number of pre-generated scenarios stored in serialized form as pickle (.pkl) files.

Algorithms are evaluated through a series of progressively more challenging tests, with increasing complexity in terms of "static" parameters. Figure 4(a) illustrates the tuning of key parameters—such as the number of agents, maximum cargo per episode, and number of airports—across all levels in the hidden set of scenarios. During the competition phase, the first scenario featured 24 agents, 72 units of cargo, and 12 airports, with no dynamic events. In contrast, the final scenario involved 426 agents, 1278 units of cargo, and 213 airports, along with a higher probability of dynamic events.

Figure 4(b) shows how the working capacity of airports decreases over time as the tests progress. At Test 1, each airport has a capacity to process 12 airplanes; however, by Test 15, this capacity drastically drops to just 1. This decrease reflects the increasing difficulty of the scenarios.

Figure 4(c) shows the growing likelihood that any route may become unavailable, while Figure 4(d**)** illustrates the duration (in steps) for which routes may remain offline. As the number of environments a competitor's algorithm must handle increases, so does the frequency and duration of route disruptions.

Finally, Figure 3(e**)** shows the rate of dynamic cargo generation at specific test levels. The rate of dynamic cargo generation is gradually increased throughout the tests and levels, introducing additional complexity as the tests advance.

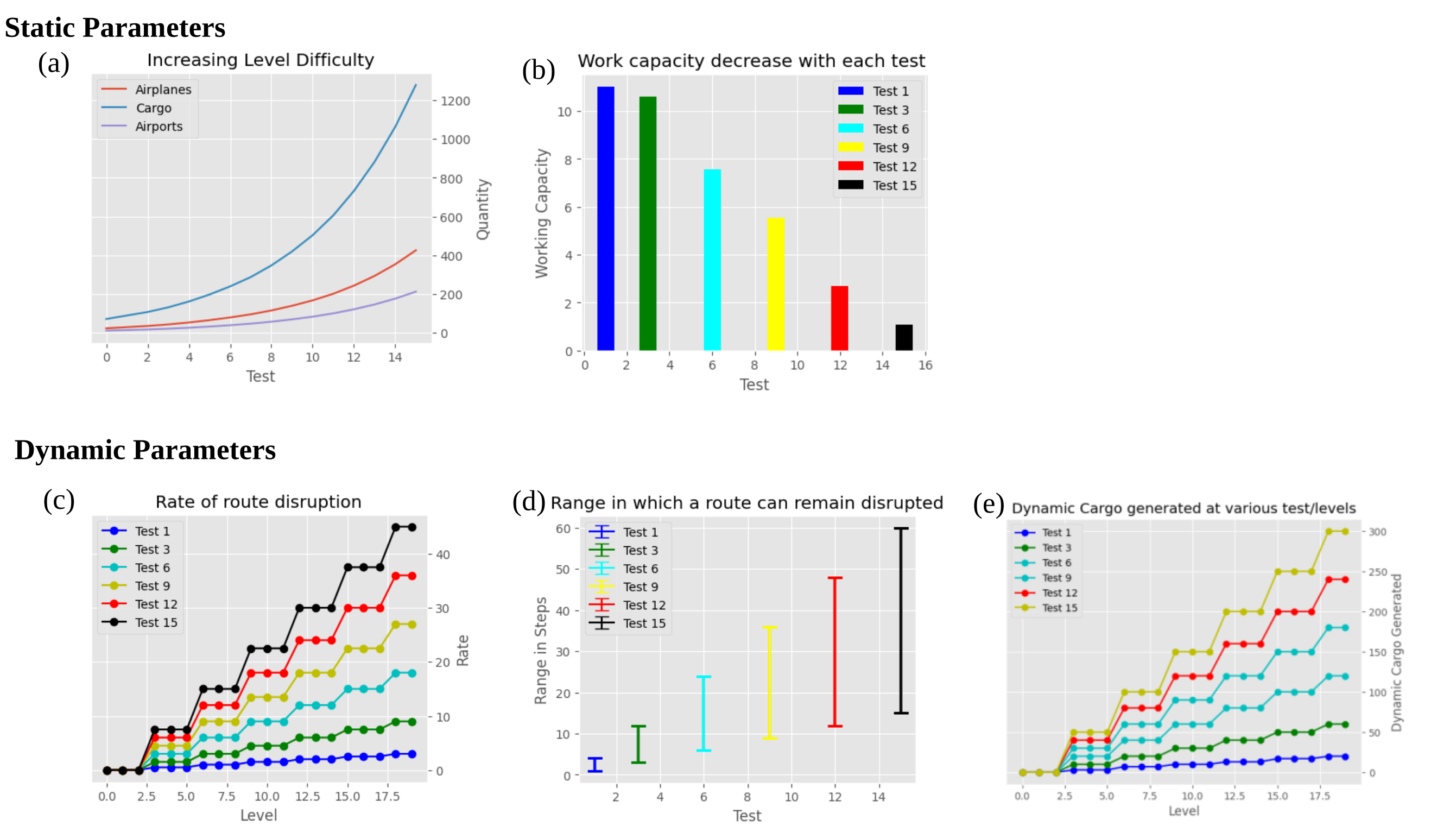

*Figure 4: Static and Dynamic parameters based on Tests and Levels*

-

[5] See https://github.com/airlift-challenge/airlift-starter-kit

## Results

The competition hosted on CodaLab began in November 2023, attracting 40 registered users. The competition was divided into Phase 1 and Phase 2. Phase 1 of the competition utilized one airplane type and assisted the team in debugging any environment issues that the participants may encounter. Phase 1 concluded on January 12th, 2024, followed immediately by the start of Phase 2. Unlike Phase 1, Phase 2 introduces two airplane types: a larger aircraft with greater carrying capacity and a smaller one with a reduced carrying capacity.

The competition concluded on April 18th, 2024 with the following winners:

- **1st place: Team JLH4E with 1201.01 points**
  Team: John F. Kolen
- **2nd place: Team FKIE with 507.70 points**
  Team: Marvin Gülhan, Jonathan Cawalla (of Fraunhofer FKIE)
- **3rd place: The Down Underdogs with 372.31 points**
  Team: Dimi Defillet, Mark Lawrenson, Prateek Aranha

Figure 5 shows the score accumulation as the evaluation progresses through each test, i.e., for Test $k$ the accumulated score is $\sum_{i=1}^{k} \sum_{j=1}^{\#\,levels} NormalizedScore(i,j)$. Team JLH4E steadily increases their lead throughout evaluation, completing a total of 12 tests before exceeding the threshold for percent of missed deliveries. Teams FKIE and The Down Underdogs both complete 10 tests before hitting this threshold. For comparison, the Baseline score is shown, which obtains a score of approximately 1 for each scenario.

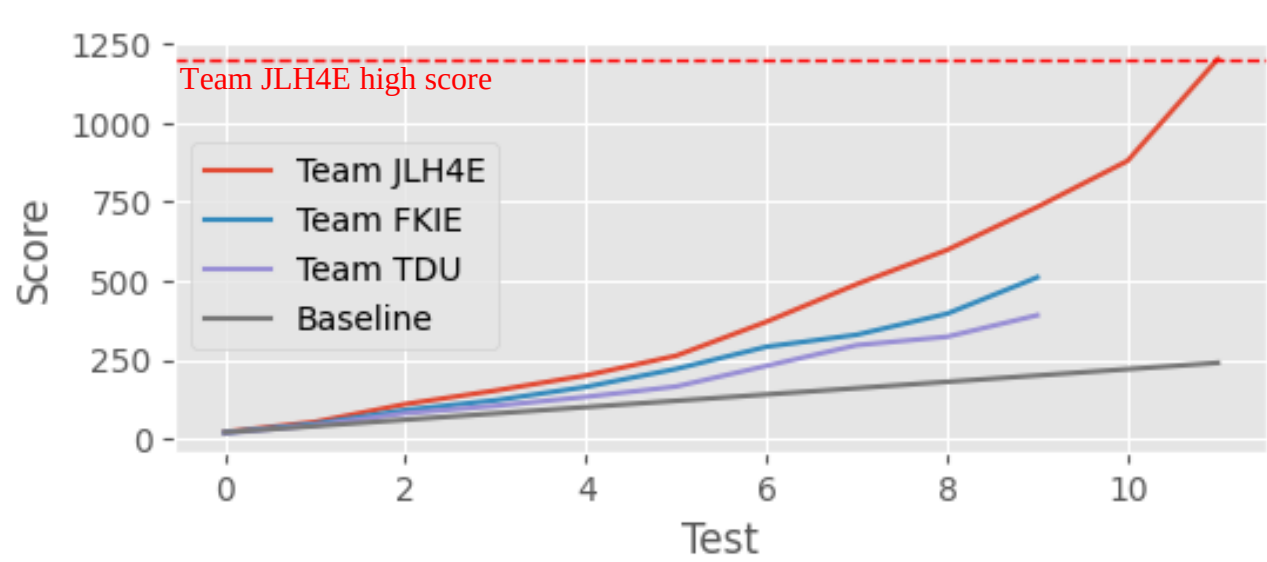

*Figure 5: Score by Test*

### **Winning Submission Approach:** Team JLH4E[6]

This section summarizes the winning solution, based on the detailed writeup by team JLH4E found in (Kolen, 2024). In the ICAPS Airlift Challenge, the team's goal was to maximize the scoring rate by solving cargo delivery tasks as quickly as possible. This "build-fast" strategy contrasted with the "build-smart" used to win the previous competition at SPIE Defense 2023. The "build-fast" strategy begins with all aircraft in a reserved state, ready to be tasked as soon as new cargo was generated. When cargo became available, the policy checked if active aircraft could pick it up and deliver it along their existing route. If no active aircraft were available, reserved aircraft would be promoted to active status to ensure timely cargo delivery. Minimizing the number of aircraft in flight helped avoid congestion at airports with limited processing capabilities.

Static map management was used to avoid recalculating routes due to disruptions (e.g., routes becoming unavailable). It was found that waiting for the route to clear was more efficient than adjusting flight paths. In many situations, there were no alternate routes; either the disruption impacted a brige edge or backtracking would take longer than waiting. The occasional reward for finding an alternative was not worth the computational cost of repetitive pathfinding. Disregarding this requirement led to faster decision-making and higher performance. In the second phase, where two aircraft types were required for cargo transport, the metamap used for route planning was never updated after its initial construction as a continuation of this strategy.

The team also offered recommendations to improve the scoring methodology, identifying two significant issues: (1) agents received scores only upon completing all levels of each test, and (2) score normalization skewed the results. Additionally, the normalized scores lacked a clear indication of the potential for improvement. They proposed that scoring could be improved by accumulating each level's score incrementally, rather than waiting for test completion, and by introducing a "zero score" baseline. This approach would offer more insights into algorithm performance, highlighting areas where further improvement is possible.

### **Winning Submission Approach:** Team FKIE[7]

*Heuristic Corridor-Based Routing for Mitigating Bottlenecks in Airlift Missions*

While examining the Airlift Challenge and the heuristic baseline method, the team identified three issues that the proposed heuristic method addresses: different plane subgraphs, bottlenecks, and malfunctions. The challenge's primary difficulty lies in the existence of two plane types, which are bound to different subgraphs. This often necessitates the involvement of up to three planes to transport cargo from the start to the destination nodes. Moreover, if too many planes land at an airport simultaneously, significant delays occur due to limited airport capacity and processing time. Additionally, routes can randomly go offline, leading to potential congestion as planes may follow the same or overlapping paths. The main idea of the method is to form corridors over the graph in a one-time planning phase and then use these corridors in an execution phase to dynamically plan flights across different subgraphs, avoid congestion, and mitigate the risk of disruption.

[6] This section includes a writeup by John F. Kolen

[7] This section was written by Marvin Gülhan and Jonathan Cawalla

In the initial planning phase, the method identifies corridors which are paths from the start to the destination nodes. These corridors may traverse multiple subgraphs. Common nodes, which exist in two subgraphs, enable cargo exchanges of different airplanes over different subgraphs. The steps to form corridors are as follows:

1. Identify common nodes in the graph.
2. Set the current subgraph as the one containing the pickup zone.
3. While the subgraph with the destination nodes is not reached: Find the shortest paths from every start node to each common node between the current and adjacent subgraphs.
4. Extend the shortest paths subgraph by subgraph.

Next, agents are assigned corresponding to the planes to each corridor, considering how many subgraphs a corridor spans. Then, one plane is selected from each subgraph that the corridor traverses, ensuring coverage for all start nodes in the pickup zone. Planes primarily follow a segment of a corridor and fly this corridor segment continuously.
During the execution phase, all agents follow these main behaviors based on their current state:

1. No cargo onboard: The agent flies towards the start of the assigned corridor segment.
2. Reached corridor start node: The agent waits until cargo becomes available at its corridor start node. This happens either because the start node is a node in the pick-up zone or because other agents deliver cargo to that node.
3. Cargo loading: The agent loads the highest priority cargo and flies towards the corridor segment's endpoint.
4. Cargo unloading: The agent unloads the cargo. If the endpoint is a target node, the agent finds a round-trip route to other target nodes before returning.

The corridor creation method inherently provides diversity in flight paths. Nevertheless, simultaneous plane arrivals at a node (due to graph characteristics) can still occur. In such cases, planes transporting cargo with the most urgent deadlines are prioritized. For malfunctions, if an alternative path within the same subgraph is shorter than waiting for the original route to become available, the route is changed.

## PDDL Domain: Pickup and Delivery with Time Windows

As a first step towards applying AI Planners to the Airlift Planning problem, the competition organizers have developed a PDDL domain that captures some core deterministic aspects of the airlift problem (Fox & Long, 2003). This model essentially embodies a Pickup and Delivery with Time Windows (PDPTW) problem (Berbeglia, Laport, & Cordeau, 2010). This new domain is referred to as the PDPTW domain.

The basic pickup and delivery problem (without time windows) has been explored through the "logistics" domain in IPC going back to the AIPS-98 Planning Competition, see (Long, et al., 2000) and (Seipp, Fawcett, Masataro, Muise, & Blonet, 2022). A domain for the closely related Vehicle Routing Problem has also been proposed in (Cheng & Gao, 2014). However, these domains do not include temporal aspects.

The PDPTW domain incorporates temporal elements in the form of durative actions (movement, loading, an unloading) and time windows via Timed Initial Literals (Haslum, Lipovetzky, Magazzeni, & Muise, 2019). This model can be considered a temporal variant of the $\text{TRANSPORT}_{\infty+}$ planning task introduced in (Helmert, 2008), which represents a transportation problem with no fuel constraints, no vehicle capacity constraints, and one vehicle type. For simplicity, the domain contains a fully connected transport graph having uniform travel times between all locations, with multiple vehicles and action costs left as a future extension.

It could be interesting to use PDDL solvers in order to compare the results between the OpenAI version and the PDDL version, or even to solve sub-components of the overall problem. There is promise in not only using other generalized planners but possibly using Large Language Models (LLMs) for planning tasks using PDDL. See for example (Silver, et al., 2023) and (Valmeekam, Sreedharan, Marquez, Olmo, & Kambhampati, 2013)

In order to more closely match the current environment, probabilistic elements of malfunctioning routes and dynamic cargo generation would have to be added. An ideal candidate for expressing this model may be RDDL (Relational Dynamic Influence Diagram Language)[8]. RDDL could also model rewards which are an important element of the OpenAI Gym interface for techniques such as reinforcement learning.

The PDPTW domain is listed in the appendix and is also available under the Airlift Challenge Github account[9].

## Future Work

Both iterations of the Airlift Challenge have brought in several high-quality submissions. Despite this success, a number of future formats are possible which may allow better or different solutions as well. Team JLH4E has several critiques, and the competition organizers are in agreement with his analysis and potential solutions. Another criticism from past participants (those interested in reinforcement learning) was that the randomized environments create entirely new air route maps for each episode, which required reinforcement learning algorithms to work in a very general manner. In fact, this was not as indicative of real-world situations where airports would stay in the same areas and the air network would remain relatively unchanged.

[8] See https://users.cecs.anu.edu.au/~ssanner/IPPC_2011/RDDL.pdf

[9] See https://github.com/airlift-challenge/PDDL-Domain

The PDDL domain may serve as an initial step towards a domain submission for a future IPC. It could allow for general planners as possible solutions for this challenge.

## Acknowledgments

The competition organizers would like to thank ICAPS for allowing them to host the Airlift Challenge competition in the conference venue, and facilitating outreach to the wider planning community.

The organizers also extend their gratitude to the Flatland challenge for its influence on the competition and code base.

Finally, they extend their appreciation and thank you to all the registered users and participants of the challenge, and most importantly, to the winners. Their contributions and bug reports played the most significant role in the success of this competition.

## Appendix: PDDL Domain listing

The PDPTW domain incorporates temporal elements from the PDDL 2.1 specification (Fox & Long, 2003). The temporal PDPTW domain is defined as follows:

```
;; Simple Pickup and Delivery with Time Windows (PDPTW) Domain

(define (domain PDPTW)
  (:requirements :strips :typing :durative-actions :timed-initial-literals)
  (:types package
          vehicle - physobj
          place
          physobj - object)
  (:predicates
    (objat ?obj - physobj ?loc - place)
    (in ?pkg - package ?veh - vehicle)
    (available ?pkg - package))

 (:durative-action LOAD
  :parameters   (?pkg - package ?vehicle - vehicle ?loc - place)
  :duration (= ?duration 1)
  :condition (and (at start (objat ?pkg ?loc))
                  (over all (and (objat ?vehicle ?loc)
                                 (available ?pkg))))
  :effect (and (at start (not (objat ?pkg ?loc)))
               (at end (in ?pkg ?vehicle))))

 (:durative-action UNLOAD
  :parameters   (?pkg - package ?vehicle - vehicle ?loc - place)
  :duration (= ?duration 1)
  :condition (and (at start (in ?pkg ?vehicle))
                  (over all (and (objat ?vehicle ?loc)
                                 (available ?pkg))))
  :effect (and (at start (not (in ?pkg ?vehicle)))
               (at end (objat ?pkg ?loc))))

 (:durative-action MOVE
  :parameters (?vehicle - vehicle ?loc-from - place ?loc-to - place)
  :duration (= ?duration 2)
  :condition (and (at start (objat ?vehicle ?loc-from)))
  :effect (and (at start (not (objat ?vehicle ?loc-from)))
               (at end (objat ?vehicle ?loc-to))))
)
```

An example problem is as follows (this problem contains three locations, three packages, and two vehicles):

```
(define (problem PDPTW-3-3-2)
(:domain PDPTW)
(:objects
 veh1 veh2 - vehicle
 loc1 loc2 loc3 - place
 pkg1 pkg2 pkg3 - package)

(:init
 ; Vehicle locations
 (objat veh1 loc1)
 (objat veh2 loc1)
 ; Package pickup locations
 (objat pkg1 loc1)
 (objat pkg2 loc2)
 (objat pkg3 loc3)
 ; Package times available for pickup
 (at 10 (available pkg1))
 (at 1 (available pkg2))
 (at 2 (available pkg3))
 ; Package delivery deadlines
 (at 15 (not (available pkg1)))
 (at 10 (not (available pkg2)))
 (at 7 (not (available pkg3)))
)

; Package delivery locations
(:goal (and  (objat pkg1 loc2)
             (objat pkg2 loc3)
             (objat pkg3 loc1)
       )
)
)
```